\documentclass[10pt,twocolumn,letterpaper]{article}

\usepackage{cvpr}
\usepackage{times}
\usepackage{epsfig}
\usepackage{graphicx}
\usepackage{amsmath}
\usepackage{amssymb}
\usepackage{ctable}
\usepackage{comment}
\usepackage[breaklinks=true,bookmarks=false]{hyperref}

\cvprfinalcopy

\begin{document}

\title{Handling Inter-Annotator Agreement for\\Automated Skin Lesion Segmentation}

\author{Vinicius Ribeiro\textsuperscript{1} ~~~Sandra Avila\textsuperscript{2} ~~~Eduardo Valle\textsuperscript{1}\\
\textsuperscript{1}School of Electrical and Computing Engineering (FEEC) ~~~\textsuperscript{2}Institute of Computing (IC)\\ 
RECOD Lab., University of Campinas (UNICAMP), Brazil\\
}

\maketitle

\begin{abstract}
In this work, we explore the issue of the inter-annotator agreement for training and evaluating automated segmentation of skin lesions. We explore what different degrees of agreement represent, and how they affect different use cases for segmentation. We also evaluate how conditioning the ground truths using different (but very simple) algorithms may help to enhance agreement and may be appropriate for some use cases. The segmentation of skin lesions is a cornerstone task for automated skin lesion analysis, useful both as an end-result to locate/detect the lesions and as an ancillary task for lesion classification. Lesion segmentation, however, is a very challenging task, due not only to the challenge of image segmentation itself but also to the difficulty in obtaining properly annotated data. Detecting accurately the borders of lesions is challenging even for trained humans, since, for many lesions, those borders are fuzzy and ill-defined. Using lesions and annotations from the ISIC Archive, we estimate inter-annotator agreement for skin-lesion segmentation and propose several simple procedures that may help to improve inter-annotator agreement if used to condition the ground truths.
\end{abstract}

\section{Introduction and Existing Art}\label{sec:introduction}

In this work, we explore the issue of the inter-annotator agreement for training and evaluating automated segmentation of skin lesions. We explore what different degrees of agreement represent, and how they affect different use cases for segmentation. We also evaluate how conditioning the ground truths using different (but very simple) algorithms may help to enhance agreement and may be appropriate for some use cases. 

Our experiments are conducted on the ISIC Archive — the largest public dataset of skin lesion images accompanied of reference segmentation by humans — and as far as we know, the only one to provide more than one reference segmentation per image. The ISIC Archive is the baseline for most of the research in the area~\cite{hu2018deep, xue2018adversarial}.

The segmentation of skin lesions is a cornerstone task for automated skin lesion analysis, useful both as an end-result to locate/detect the lesions and as an ancillary task for lesion classification. Lesion segmentation, however, is a very challenging task, due not only to the challenge of image segmentation itself but also to the difficulty in obtaining properly annotated data. Detecting accurately the borders of lesions is challenging even for trained humans, since, for many lesions, those borders are fuzzy and ill-defined.

Since the inception of automated skin lesion analysis, the segmentation of lesions has attracted scientific interest~\cite{dhawan1992segmentation, moss1996unsupervised}. Early methods of lesion classification tended to strictly mimic medical criteria~\cite{fornaciali2016towards}, such as the ABCD rule~\cite{nachbar1994abcd}, in which both (B)order irregularity and large (D)iameter depend on lesion segmentation to be estimated automatically. Such methods were also consonant with the art on computer vision of the 1990s, in which segmentation was often considered a crucial preliminary step for classification (e.g., to allow extracting shape features).

The transition of computer vision art to bags-of-words models in the 2000s ~\cite{sivic2006video}, and to deep learning in the 2010s~\cite{krizhevsky2012imagenet} spelled the end of the viewpoint of segmentation as an ancillary technique in preparation for classification. That understanding, however, also increased the appreciation of segmentation for its own merits. With the accumulated experience brought by collective efforts like the PASCAL VOC~\cite{everingham2010pascal} and the ImageNet~\cite{deng2009imagenet} challenges, we now understand not only that segmentation and classification can be tackled independently, but also that segmentation is usually \textit{much more challenging} than classification.

Those advances in computer vision appear in the current art in skin lesion analysis~\cite{fornaciali2016towards, menegola2017knowledge, bissoto2018deep, perez2018data}, in which, although lesion segmentation is sometimes still used to help in the classification, it is largely understood as an important and challenging task in itself.

Obtaining accurate annotations is paramount for all machine learning techniques. The accuracy of annotations imposes an upper bound on the \textit{actual, real world} accuracy of learned models. Although, in theory, any model can reach 100\% of accuracy on any dataset, accuracies above those of the annotations only reflect the ability of models of learning the datasets' biases. Thus, appraising annotation accuracy is important to decide the point above which it becomes counterproductive to keep working on the models. Estimating annotation accuracy is often, however, impossible, since it requires, in principle a \textit{more reliable} standard than the one provided by the annotations themselves. In scenarios where such a standard is not available, the inter-annotator agreement can act as a proxy estimation.

Segmentation tasks, especially, bring challenges for annotation accuracy, due both to the more complex procedure of annotating borders and regions (in comparison to just providing a label), and to the often subjective nature of the task, in which the position of a border/limits of a region may be ill-defined --- in particular for some skin lesions, with low contrast and very fuzzy borders (Figure~\ref{fig:high_variability}).

Existing art on the inter-annotator agreement for segmentation is very scarce. Contrarily to existing works for lesion classification~\cite{esteva2017dermatologist, brinker2019comparing, haenssle2018man}, we could not find any evaluation of annotator accuracy or inter-annotator agreement for skin-lesion segmentation. Even for other tasks in medical images, systematic studies of the inter-annotated agreement are hard to find. 

The most complete study we found was by Lampert \etal~\cite{lampert2016empirical}, who presents an in-depth study of the inter-annotator agreement for four image processing problems --- segmentation of natural images, fissures in remote-sense images, landslides in satellite images, and blood vessels in retinoscopy --- employing a large number of analytic tools to explore agreement on those tasks. The most relevant (and easy to interpret) result is the one that compares the performance of each annotator with the consensus annotation (obtained averaging the annotations). For the retinoscopy task, they had only two annotators, with Cohen's Kappa scores of 0.50 and 0.57 when compared to consensus. 

Liedlgruber \etal~\cite{liedlgruber2016hippo} evaluate the segmentation of the hippocampus in Magnetic Resonance Image volumes for 9 patients, by 3 different annotators, who used a graphic table to delineate the hippocampus voxels on each slice of the image. They report large variations of agreements between the three pairs of annotators and across the 9 patients, with an average 76\% agreement using the Dice score, and 6.5 using the Symmetric Hausdorff distance.

Chaichulee \etal~\cite{chaichulee2017multi} report results for segmentation of areas of exposed skin on patients, aiming at non-contact vital signal monitoring. On a dataset comprising over 200 hours of video acquired from the recording of 15 pre-term infants in intensive neonatal care, they asked 3 annotators to label the regions of exposed skin, in a semi-automated procedure where the annotator would annotate one frame, the system would attempt to propagate the annotation for the next frames, and the annotators would accept, or revise the propagation. They report a mean agreement of 96.54\% using the Jaccard index and also provide an estimation of the distribution of the agreements in the form of a histogram. 

An extended abstract by Egger \etal~\cite{egger2016clinical} presents results for mandibular bone segmentation on high-resolution (512$\times$512) 3D Computer Tomography scans. They asked two specialists to annotate the datasets and measured an agreement of 93.67\% using the Dice score.

The results above suggest that inter-annotator agreement for segmentation may vary widely, according to the nature of the image, and the details of the task.

An important consideration to appraise the impact of annotation accuracy for segmentation is its indented use. For skin lesions, we can easily identify at least three very distinct use cases, with progressively stricter demands of accuracy:

\noindent\textbf{Localization:} here we are interested in \textit{detecting} the presence of the lesions, and \textit{locating} its position. The precise limits of the lesion are not important. For this use case, an approximate bounding box may suffice, or even less: a single point anywhere inside the lesion may be enough. This level of annotation may be useful, for example, for automatically locating the lesions in a full-body skin exam.

\noindent\textbf{Demarcation:} here we must not only locate the lesions but also correctly determine their \textit{overall shape}. We want to be able to estimate metrics such as the lesion diameter, eccentricity, and overall symmetry. 

\noindent\textbf{Description:} here we want to fully characterize the lesion border, including detailed characteristics such as smoothness vs. irregularity. This level of annotation is the one required to mimic the medical algorithms (e.g., the ABCD rule) in a straightforward way. 

The list above does not intend to be exhaustive, just to illustrate how different use cases may impose very different demands to both the ground-truth annotators and to the automated techniques.

In this work, we will discuss the impact of different levels of inter-annotator agreement on those use cases, and explore how very simple conditionings may significantly improve the agreement for some use cases. Our main contributions are:

\begin{itemize}
  \item An estimation of the inter-annotator agreement for skin-lesion segmentation. We not only provide simple statistics (such as a mean) but instead attempt to fully characterize the distribution of the agreements;
  \item A visual presentation of representative samples for different agreements, in order to help the reader to grasp their qualitative meaning;
  \item An evaluation of several simple procedures that may help to improve inter-annotator agreement if used to condition the ground truths. Those conditionings may be helpful for some use cases.
\end{itemize}

In this paper, we do not propose, neither evaluate segmentation techniques \textit{per se}: our whole focus is on the data. All the code used to condition the ground truths and analyze the results is available.

We have already reviewed the little existing literature on the inter-annotated agreement for medical images in this introduction. The remainder of this paper is organized as follows. In Section~\ref{sec:materials_methods}, we first describe the most important dataset in current research on skin lesion analysis, we then describe the statistics used to evaluate the inter-annotator agreement and the conditionings we explore. In Section~\ref{sec:experimental_design}, we describe the experiments we run and the basis of our statistics. In Sections~\ref{sec:results}, we present our results. We present the statistics for the data under evaluation and we explain the different regions of the statistical distributions of inter-annotator agreement. Finally, we draw our conclusions in Section~\ref{sec:conclusions}.

\begin{figure*}
\begin{center}
\includegraphics[height=0.9\textheight]{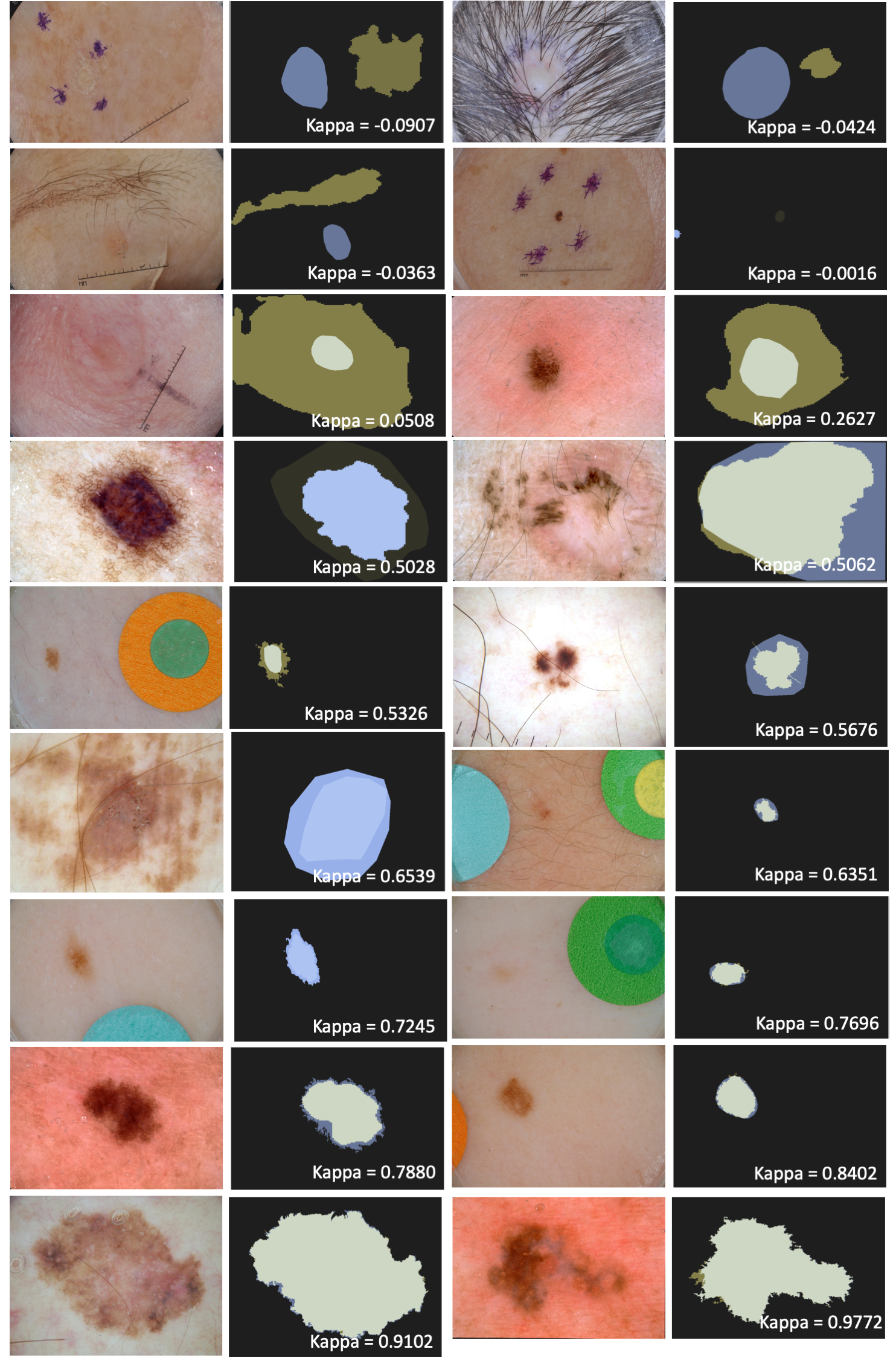}
\end{center}
    \caption{Samples extracted from the International Skin Imaging Collaboration (ISIC) Archive dataset. We have samples representative of all inter-annotator agreements we found in the distributions we observed. In our study, each skin lesion has at least two segmentation ground truths. The inter-annotator agreement is worse than random when the Kappa score is below 0. Kappa scores above 0.8 are considered high. The examples here may help the reader to appreciate qualitatively the meaning of different scores.}
\label{fig:low_agreement}
\end{figure*}

\section{Materials and Methods}\label{sec:materials_methods}

\subsection{Dataset}\label{sec:dataset}

This work is based upon the ISIC Archive~\cite{isicarchive} --- curated by the International Skin Imaging Collaboration --- the largest publicly available dataset of images of skin lesions, with over 23\,000 annotated images. Although a few other datasets also provide segmentation information~\cite{ballerini2013color, mendonca2015ph2}, as far as we know, the ISIC Archive is the only public dataset with more than one segmentation annotation per lesion, and thus the only one where inter-annotator agreement can be appraised.

At the time we ran our experiments, the ISIC Archive dataset contained exactly 23\,907 images of lesions, 13\,779 of which had segmentation ground truths. For our study, however, we need images with at least \textit{two} ground truths, reducing those to the much smaller subset of 2\,233. 

A subset of the ISIC Archive was employed on the ISIC Challenges, which included a task for lesion segmentation~\cite{isic2016, isic2017, isic2018}. Since the challenge allowed for the first time the researchers to directly compare their techniques in a fully reproducible setting, it has been very influential in the community. Therefore, in addition to analyzing the full archive, we also explored the image subsets used on the past two challenges, to see if there were any appreciable differences. Table~\ref{table:annotations_distribution} summarizes all three datasets.

\begin{figure}
\begin{center}
\includegraphics[width=0.9\linewidth]{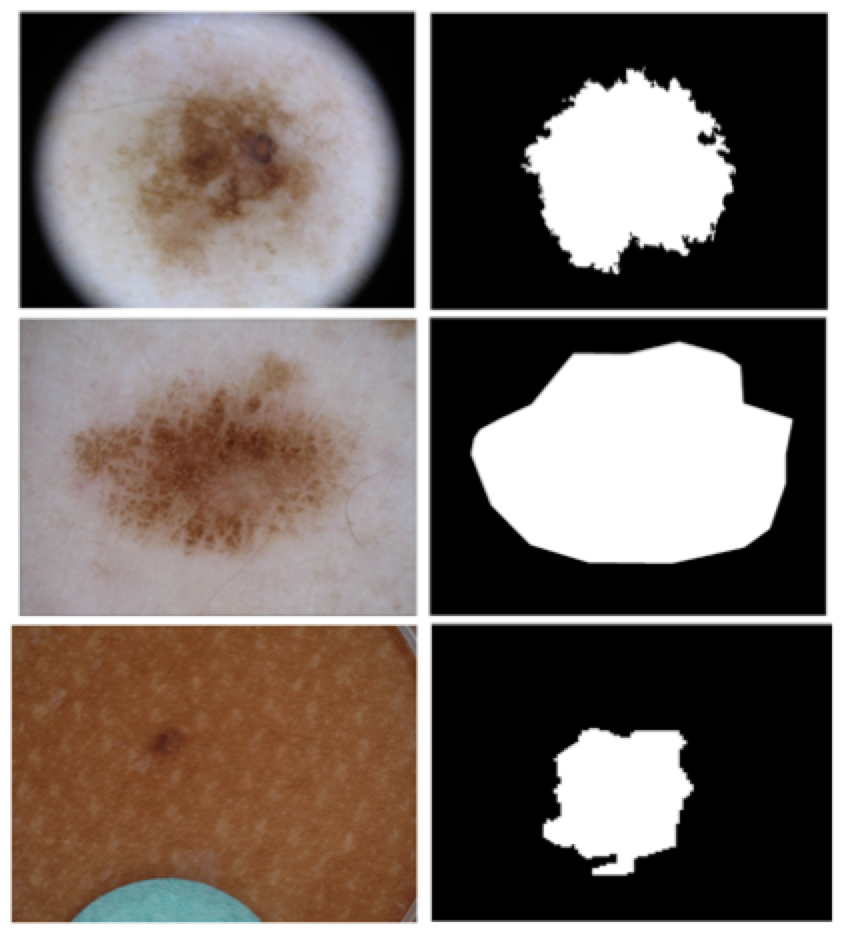}
\end{center}
   \caption{Samples extracted from the ISIC Archive dataset. Top: flood-fill algorithm controlled by the annotator. Middle: manual polygon tracing. Bottom: fully-automated annotation validated by a human annotator.}
\label{fig:high_variability}
\end{figure}

The ground truth annotations in the ISIC Archive are highly variable. In fact, just considering the subsets used for the ISIC Challenges, there are already three different methods to create the annotations. As stated by the Challenge organizers~\cite{isic2018_task1}: (1) a semi-automated flood-fill algorithm, with parameters chosen by a human expert; (2) a manual polygon tracing by a human expert; (3) a fully-automated algorithm, reviewed and accepted by a human expert. As shown in Figure~\ref{fig:high_variability}, the first method tends to create a very irregular border, the second very smooth borders, and the third is in-between, with borders that appear ``pixelated''.

\begin{table}[t]
\begin{center}
\begin{tabular}{cccc}
\toprule
Annotations & ISIC Archive & ISIC 2017 & ISIC 2018 \vspace{.05cm} \\
\toprule
1 & 11\,546 & 1\,290 & 1\,488 \\
2 & 2\,094 & 616 & 995 \\
3 & 100 & 67 & 71 \\
4+ & 39 & 27 & 34 \\
\toprule
Total & 13\,779 & 2\,000 & 2\,588 \vspace{.05cm}\\
\toprule
\end{tabular}
\end{center}
\caption{Number of available annotations per image for each dataset.}
\label{table:annotations_distribution}
\end{table}

\subsection{Methods}\label{sec:methods}

In this work, we not only measure the inter-annotator agreement on the original ground truths but also evaluate how simple conditioning of the ground truths may help to enhance that agreement.

The conditioning consists of applying simple image processing operations to all ground truth masks. The proposed conditionings are very straightforward and deterministic --- there is no learning involved. We list them below:
 
\begin{itemize}
  \item Opening: this is a morphological operation that removes details from the foreground (lesion). The structuring element was a square of five pixels; 
  \item Closing: this is a morphological operation that removes details from the \textit{background}, e.g., small holes or tears. Same structuring element as above;
  \item Convex hull: here we find the smallest convex shape that covers the entire lesion;
  \item Opening or Closing + Convex hull: the morphological operation followed by the convex hull;  
  \item Bounding box: here we find the smallest rectangle with sides parallel to the image that covers the entire lesion.
\end{itemize}

\begin{figure*}
\begin{center}
\includegraphics[width=\linewidth]{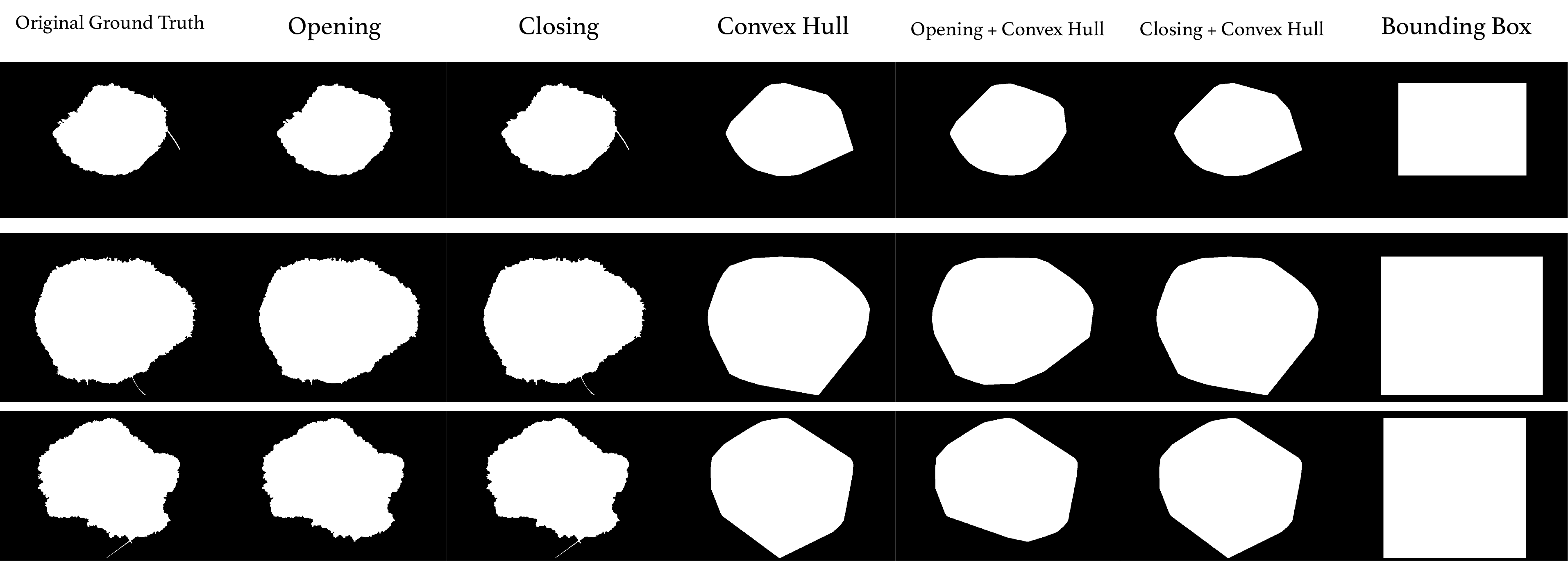}
\end{center}
   \caption{Samples extracted from the ISIC Archive dataset. For each sample, we present its corresponding mask conditioned with opening, closing, convex hull, convex hull after opening, convex hull after closing, and bounding box. Note how the opening is able to remove small details from the foreground, which may greatly affect the convex hull.}
\label{fig:transforms}
\end{figure*}

Figure~\ref{fig:transforms} illustrates those operations. From a theoretical point of view, the conditioning may be interpreted as \textbf{denoising} operations, whose aim is to preserve the cogent information about the lesion segmentation, while discarding details which depend on the choice of one particular annotator.

We implemented all the conditionings in Python. Apart from the bounding box, which was developed from scratch by our team, all the conditionings and structuring elements were extracted from the morphology package of the scikit-image library~\cite{skimage}. Auxiliary code was developed using the numpy library~\cite{numpy}. The code we used to both condition the ground truths and to analyze the results is available at our Github repository\footnote {https://github.com/vribeiro1/skin-lesion-segmentation-agreement}.

\section{Experimental Design}\label{sec:experimental_design}

As a metric for agreement, we employ the Cohen's Kappa score~\cite{mchugh2012interrater}, which offers, over the alternatives, the advantage of taking into account the probability of the agreement occurring by chance. The score ranges from $-1$ to $1$, is zero for pure chance, positive for better than chance, and negative for worse than chance. 

For a given lesion, we compute the kappa score between its ground truth annotations. If a lesion has more than two ground truths, we take the average of the kappa of all possible pairs. We tabulate all kappas to estimate the distribution of the values (and associated statistics) for a given dataset. 

To evaluate the impact of the proposed conditionings, we apply them, by turn to the ground truths before computing the scores and estimating the distributions. We employ the Kolmogorov–Smirnoff (K–S) test to check which pairs of distributions are significantly different.

\section{Results}\label{sec:results}

\begin{figure*}[!h]
\begin{center}
\includegraphics[height=0.45\textheight]{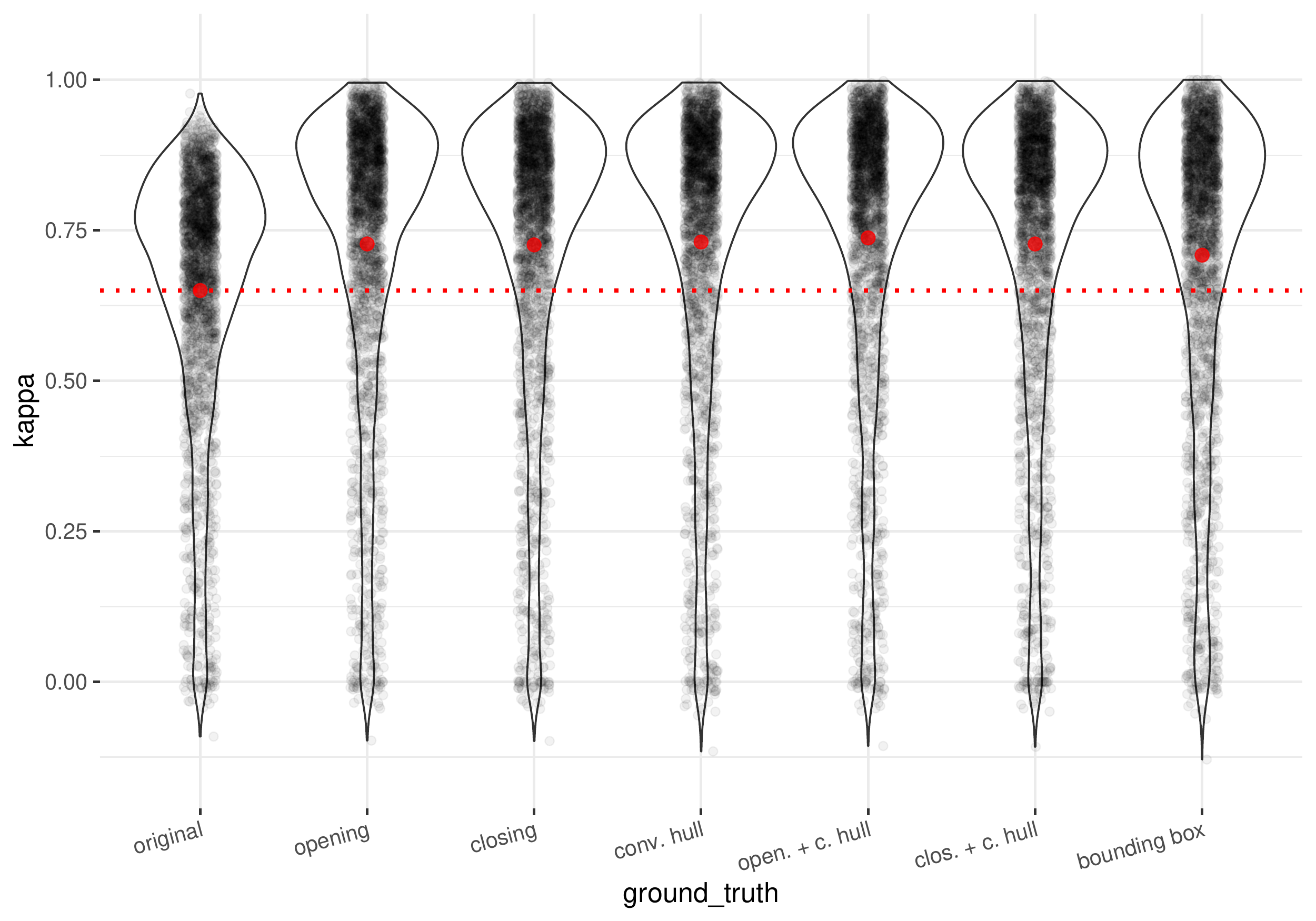}
\includegraphics[height=0.45\textheight]{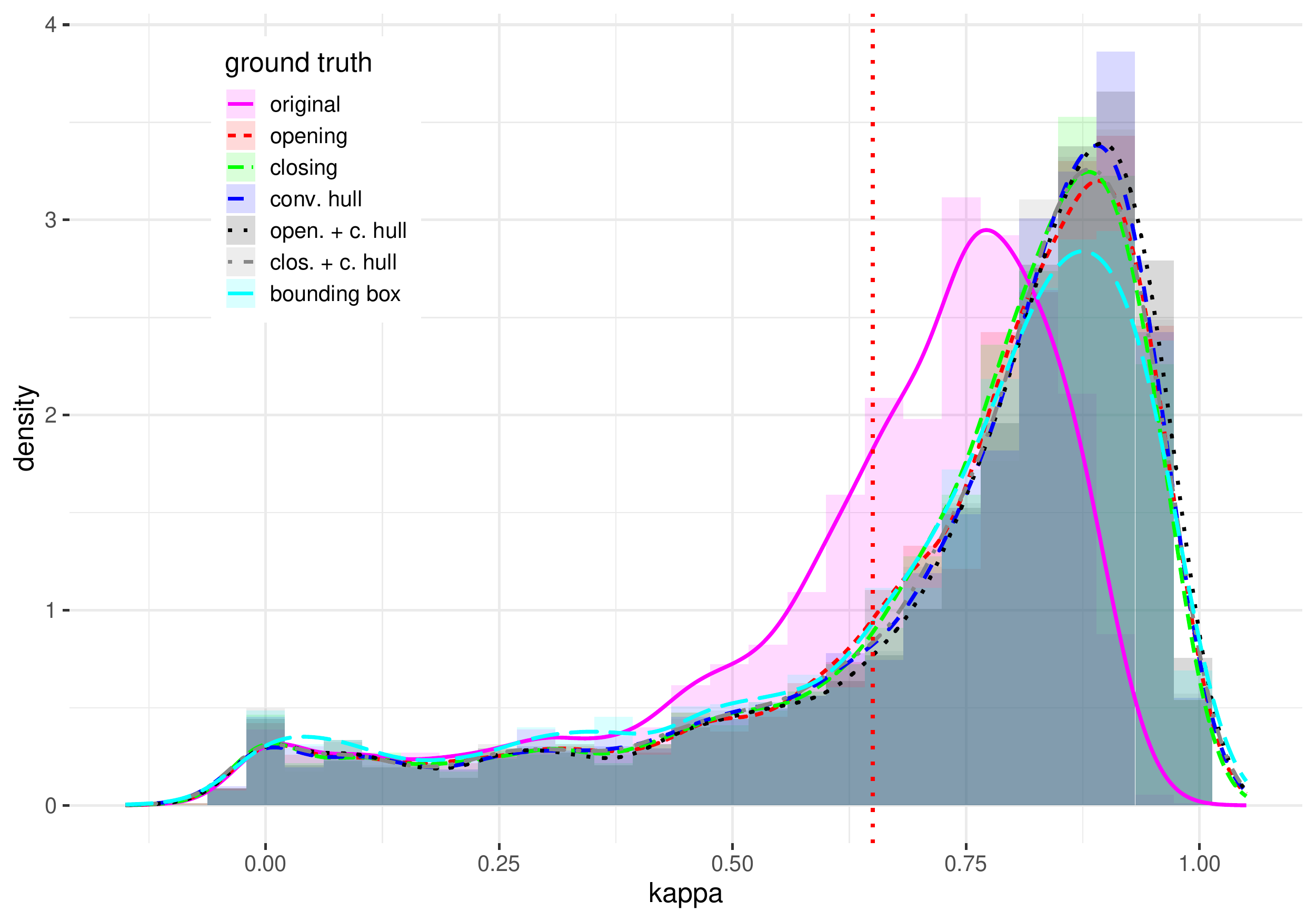}
\end{center}
   \caption{Distributions of inter-annotator agreements for the ground truths pre- (original) and post- the proposed conditionings (others). Both plots show exactly the same data. The bottom graph has the histograms (shaded areas) and the estimated densities (superimposed lines). The top graph has the original samples (black dots), the estimated densities (violin plots), and the estimated means (red dot) for each distribution. The plots show the data for the ISIC Archive.}
\label{fig:isic_archive_analysis}
\end{figure*}

\begin{figure*}[!h]
\begin{center}
\includegraphics[height=0.45\textheight]{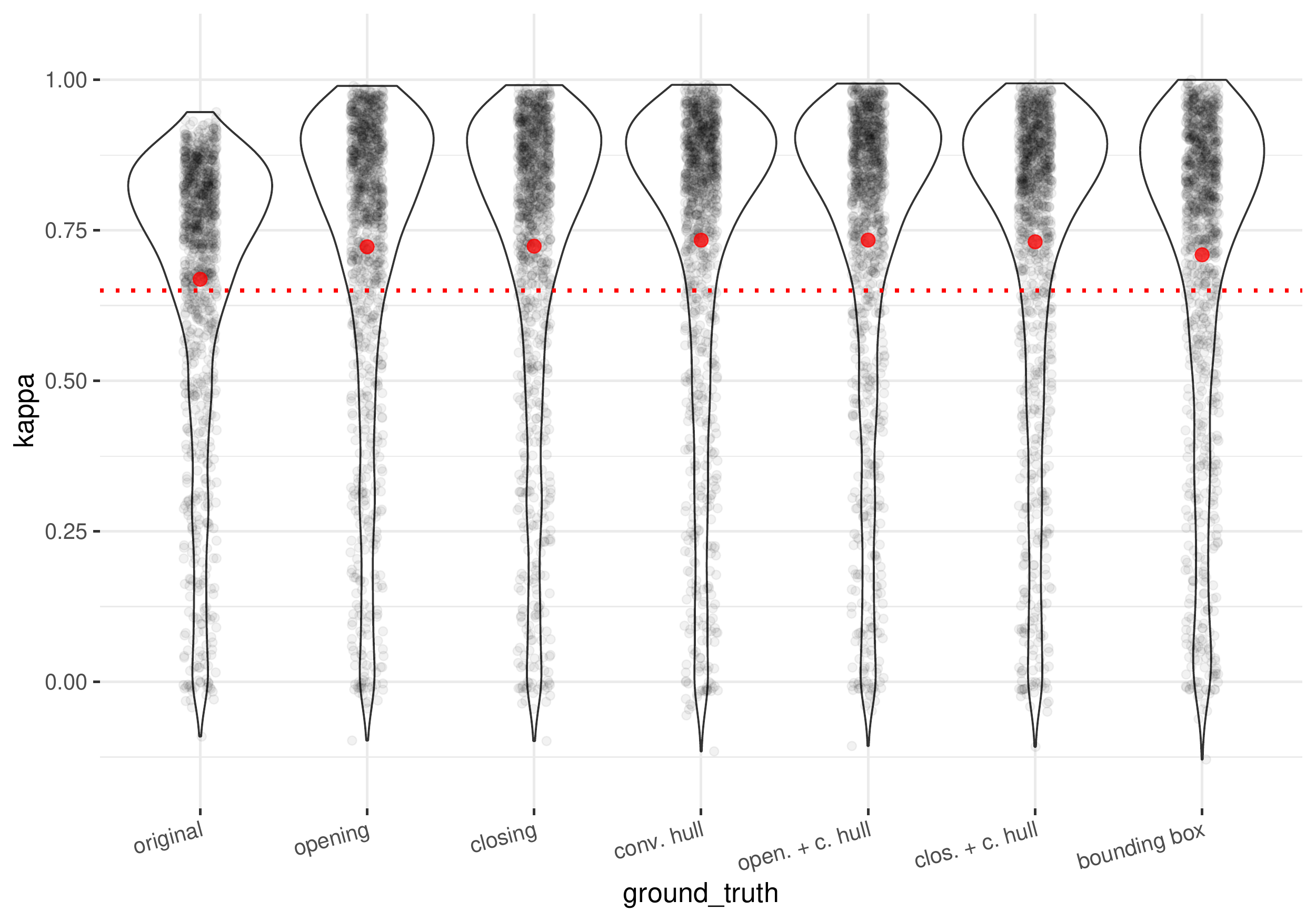}
\includegraphics[height=0.45\textheight]{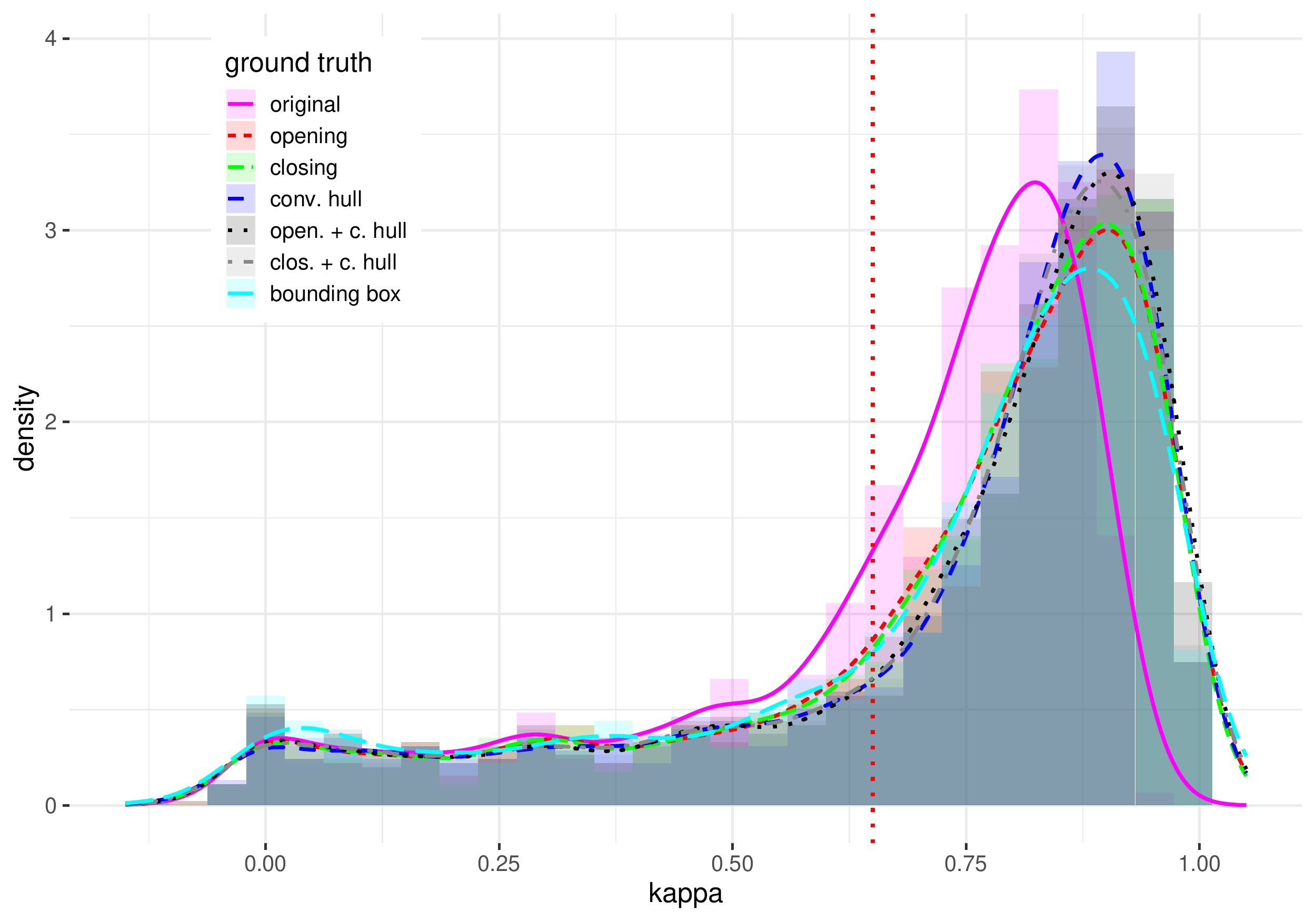}
\end{center}
   \caption{Distributions of inter-annotator agreements for the ground truths of the subset of the ISIC Archive used on the ISIC Challenge 2018. Please see Figure~4 and Section~4 for an explanation. The plots for the ISIC Challenge 2017 were essentially identical and were omitted for brevity. Note how all proposed conditionings allow improving inter-annotator agreement both here and on Figure~4.}
\label{fig:isic_2018_analysis}
\end{figure*}

The distributions of the kappa scores observed, for the original ground truths, and for all proposed conditionings, appear in Figure~\ref{fig:isic_archive_analysis} for the ISIC Archive and in Figure~\ref{fig:isic_2018_analysis} for the subset used on the ISIC 2018 Challenge. The plots for the ISIC 2017 resulted essentially identical to those for ISIC 2018 and thus were omitted for brevity. 

The upper and lower parts of the figures plot the same information in a different form. The bottom part is perhaps more straightforward to interpret: shaded areas are the (normalized) histograms of the observed Cohen kappa scores, and the line plots superimposed to them are the distributions estimated with a kernel density estimation. The upper part is more challenging to interpret but has the advantage to be much less crowded. In it, each experiment appears separated: the dots represent the actual observations (with a small random horizontal jitter to help the visualization), the shapes around each group of points (violin plots) are the distributions estimated with a kernel density estimation (the shapes are wider where the distributions are denser), and the large red dots are the means of the distributions. The values of the scores for the quantiles of the original distributions are in Table~\ref{table:distributions_percentiles} presents statistics for each dataset.

The distributions are highly skewed, with a strong mode towards high scores but a very long tail towards very low scores. The most interesting result is that all conditionings improved the ``good'' mode considerably and that most of them are indistinguishable from each other in terms of that improvement. That is surprising since the morphological conditionings (opening and closing) are much more conservative than the convex hull, but all for treatments combining those three operations obtained essentially the same results. Also surprising was that use of the bounding box --- a much more destructive choice --- was slightly \textit{worse} than the other options. 

None of the methods was able to improve the very divergent cases at the tail of the distribution: that was not unexpected since the small adjustments they make are not meant to reconcile those extreme cases. On the other hand, with the exception of the bounding box, the techniques neither worsened the tail, which was a good outcome.

There is a small difference between the application of the convex hull and the use of the morphological operators alone, but we could not show that this difference is statistically significant. The K–S test rejected the equivalence of the original distribution with all conditionings, with tiny p-values (all p-values $<10^{-20}$). It failed to reject most of the other pairs, with the notable exception of the bounding box vs. all conditionings with the convex hull ($10^{-7}<$ p-value $<0.002$).

\begin{table}[t]
\begin{center}
\begin{tabular}{cccc}
\toprule
Percentile & ISIC 2017 & ISIC 2018 & ISIC Archive \vspace{0.05cm}\\
\toprule
25\% & 0.5724 & 0.5991 & 0.5748 \\
50\% & 0.7438 & 0.7552 & 0.7185 \\
75\% & 0.8213 & 0.8312 & 0.8010 \\
95\% & 0.8838 & 0.8952 & 0.8812 \vspace{0.05cm}\\
\toprule
\end{tabular}
\end{center}
\caption{Percentiles of the Cohen's Kappa mean score distributions for each dataset.}
\label{table:distributions_percentiles}
\end{table}

\section{Discussion}\label{sec:conclusions}

Image segmentation is among the areas of computer vision that most advanced in recent years. Not only the techniques have improved sharply, but our understanding of the role of segmentation in the recognition pipeline, as well as its relationship with the task of classification,  have changed drastically. However, obtaining properly annotated data to train and evaluate segmentation models continues to be a challenge. While datasets for (general) image classification have now millions of samples and thousands of categories, segmentation datasets are considerably smaller. For medical images, annotated data for segmentation is even scarcer.

Our results demonstrate the challenge of annotating skin lesions, showing that the median inter-annotation agreement for humans has around 0.72 kappa score for the whole ISIC Archive and slightly more than that (~0.75) for the images selected for the challenges. The good news is that very simple image-processing techniques may significantly improve those agreements, without modifying too much the ground truths. Different applications may choose different conditionings according to their use cases: for location or demarcation, the convex hull may be the best, while for description the morphological operators, which preserve most of the border characteristics may be the best. An interesting result we found is that the strong simplification brought by the bounding boxes worsened the annotation agreements in comparison to the other techniques.

From a theoretical point of view, the conditioning may be interpreted as denoising operations, whose aim is to preserve the cogent information about the lesion segmentation, while discarding details which depend on the choice of one particular annotator. Therefore they may help both to train more robust machine learning models and to evaluate them more fairly.

The bad news is that none of the conditioning is able to deal with strong divergences. Our results show that, although most masks have a reasonable-to-good inter-annotator agreement, there is a non-negligible tail of very disparaging annotations both in the ISIC Archive as a whole and on the subsets used on the challenges. That tail, and the difficulty in deciding which of the alternative annotations is the right one might explain why during the most recent challenge of 2018, none of the five top-ranked participants of the lesion boundary segmentation employed extra data for training (from the Archive, for example), while the four top-ranked participants for lesion classification (diagnosis) employed extra data. 

In follow-up work, we would like to evaluate exactly how our conditionings impact the research on machine learning models, by attempting to measure their effect on the training and evaluation of those models. Such evaluation is far from obvious since the aim is to evaluate how models trained in a given setting generalize when exposed to different situations, in order to evaluate their robustness. The ideal design would employ a cross-dataset evaluation, testing the models with images acquired and annotated under new conditions, but a possible --- more feasible --- alternative would be to use data augmentation techniques to simulate that design. 

{\small
\bibliographystyle{ieee_fullname}
\bibliography{arxiv}

\begin{thebibliography}{10}\itemsep=-1pt

\bibitem{isicarchive}
{International Skin Imaging Collaboration: Melanoma Project}.
\newblock \url{https://isic-archive.com}.

\bibitem{isic2018_task1}
{ISIC 2018 - Task 1: Lesion Boundary Detection}.
\newblock \url{https://challenge2018.isic-archive.com/task1/}.

\bibitem{numpy}
{NumPy}.
\newblock \url{http://www.numpy.org/}.

\bibitem{skimage}
{Scikit-Image: Image processing in Python}.
\newblock \url{https://scikit-image.org/}.

\bibitem{ballerini2013color}
L. Ballerini, R.~B. Fisher, B. Aldridge, and J. Rees.
\newblock A color and texture based hierarchical k-nn approach to the
  classification of non-melanoma skin lesions.
\newblock In {\em Color Medical Image Analysis}, pages 63--86. Springer, 2013.

\bibitem{bissoto2018deep}
A. Bissoto, F. Perez, V. Ribeiro, M. Fornaciali, S.a Avila, and E. Valle.
\newblock Deep-learning ensembles for skin-lesion segmentation, analysis,
  classification: Recod titans at isic challenge 2018.
\newblock {\em arXiv preprint arXiv:1808.08480}, 2018.

\bibitem{brinker2019comparing}
T.~J. Brinker, A. Hekler, A. Hauschild, C. Berking, B. Schilling, A.~H. Enk, S.
  Haferkamp, A. Karoglan, C. von Kalle, M. Weichenthal, et~al.
\newblock Comparing artificial intelligence algorithms to 157 german
  dermatologists: the melanoma classification benchmark.
\newblock {\em European Journal of Cancer}, 111:30--37, 2019.

\bibitem{chaichulee2017multi}
S. Chaichulee, M. Villarroel, J. Jorge, C. Arteta, G. Green, K. McCormick, A.
  Zisserman, and L. Tarassenko.
\newblock Multi-task convolutional neural network for patient detection and
  skin segmentation in continuous non-contact vital sign monitoring.
\newblock In {\em IEEE International Conference on Automatic Face \& Gesture
  Recognition}, pages 266--272, 2017.

\bibitem{isic2017}
N.~C.~F. Codella, D. Gutman, M.~E. Celebi, B. Helba, M.~A. Marchetti, S.~W.
  Dusza, A. Kalloo, K. Liopyris, N. Mishra, H. Kittler, et~al.
\newblock {Skin lesion analysis toward melanoma detection: A challenge at the
  2017 International Symposium on Biomedical Imaging (ISBI), hosted by the
  International Skin Imaging Collaboration (ISIC)}.
\newblock In {\em IEEE International Symposium on Biomedical Imaging}, pages
  168--172, 2018.

\bibitem{isic2018}
N.~C.~F. Codella, V. Rotemberg, P. Tschandl, M.~E. Celebi, S. Dusza, D. Gutman,
  B. Helba, A. Kalloo, K. Liopyris, M. Marchetti, et~al.
\newblock {Skin Lesion Analysis Toward Melanoma Detection 2018: A Challenge
  Hosted by the International Skin Imaging Collaboration (ISIC)}.
\newblock {\em arXiv preprint arXiv:1902.03368}, 2019.

\bibitem{deng2009imagenet}
J. Deng, W. Dong, R. Socher, L.-J. Li, K. Li, and L. Fei-Fei.
\newblock Imagenet: A large-scale hierarchical image database.
\newblock In {\em IEEE Conference on Computer Vision and Pattern Recognition},
  pages 248--255, 2009.

\bibitem{dhawan1992segmentation}
A.~P. Dhawan and A. Sim.
\newblock Segmentation of images of skin lesions using color and texture
  information of surface pigmentation.
\newblock {\em Computerized Medical Imaging and Graphics}, 16(3):163--177,
  1992.

\bibitem{egger2016clinical}
J. Egger, K. Hochegger, M. Gall, K. Reinbacher, K. Schwenzer-Zimmerer, J.
  Wallner, and D. Schmalstieg.
\newblock Clinical evaluation of mandibular bone segmentation.
\newblock {\em IEEE Engineering in Medicine and Biology Society}, 2016.

\bibitem{esteva2017dermatologist}
A. Esteva, B. Kuprel, R.~A. Novoa, J. Ko, S.~M. Swetter, H.~M. Blau, and S.
  Thrun.
\newblock Dermatologist-level classification of skin cancer with deep neural
  networks.
\newblock {\em Nature}, 542(7639):115, 2017.

\bibitem{everingham2010pascal}
M. Everingham, L. Van~Gool, C.~K.~I. Williams, J. Winn, and A. Zisserman.
\newblock The pascal visual object classes (voc) challenge.
\newblock {\em International Journal of Computer Vision}, 88(2):303--338, 2010.

\bibitem{fornaciali2016towards}
M. Fornaciali, M. Carvalho, F.~V. Bittencourt, S. Avila, and E. Valle.
\newblock Towards automated melanoma screening: Proper computer vision \&
  reliable results.
\newblock {\em arXiv preprint arXiv:1604.04024}, 2016.

\bibitem{haenssle2018man}
H.~A. Haenssle, C. Fink, R. Schneiderbauer, F. Toberer, T. Buhl, A. Blum, A.
  Kalloo, A.~Ben~Hadj Hassen, L. Thomas, A. Enk, et~al.
\newblock Man against machine: diagnostic performance of a deep learning
  convolutional neural network for dermoscopic melanoma recognition in
  comparison to 58 dermatologists.
\newblock {\em Annals of Oncology}, 29(8):1836--1842, 2018.

\bibitem{hu2018deep}
Z. Hu, J. Tang, Z. Wang, K. Zhang, L. Zhang, and Q. Sun.
\newblock Deep learning for image-based cancer detection and diagnosis- a
  survey.
\newblock {\em Pattern Recognition}, 83:134--149, 2018.

\bibitem{krizhevsky2012imagenet}
A. Krizhevsky, I. Sutskever, and G.~E. Hinton.
\newblock Imagenet classification with deep convolutional neural networks.
\newblock In {\em Advances in Neural Information Processing Systems}, pages
  1097--1105, 2012.

\bibitem{lampert2016empirical}
T.~A. Lampert, A. Stumpf, and P. Gan{\c{c}}arski.
\newblock An empirical study into annotator agreement, ground truth estimation,
  and algorithm evaluation.
\newblock {\em IEEE Transactions on Image Processing}, 25(6):2557--2572, 2016.

\bibitem{liedlgruber2016hippo}
M. {Liedlgruber}, K. {Butz}, Y. {Höller}, G. {Kuchukhidze}, A. {Taylor}, O.
  {Tomasi}, E. {Trinka}, and A. {Uhl}.
\newblock Variability issues in automated hippocampal segmentation: A study on
  out-of-the-box software and multi-rater ground truth.
\newblock In {\em 2016 IEEE 29th International Symposium on Computer-Based
  Medical Systems (CBMS)}, pages 191--196, June 2016.

\bibitem{isic2016}
M.~A. Marchetti, N.~C.~F. Codella, S.~W. Dusza, D.~A. Gutman, B. Helba, A.
  Kalloo, N. Mishra, C. Carrera, et~al.
\newblock {Results of the 2016 International Skin Imaging Collaboration
  International Symposium on Biomedical Imaging challenge: Comparison of the
  accuracy of computer algorithms to dermatologists for the diagnosis of
  melanoma from dermoscopic images}.
\newblock {\em Journal of the American Academy of Dermatology}, 78(2):270--277,
  2018.

\bibitem{mchugh2012interrater}
M.~L. McHugh.
\newblock Interrater reliability: the kappa statistic.
\newblock {\em Biochemia medica: Biochemia medica}, 22(3):276--282, 2012.

\bibitem{mendonca2015ph2}
T.~F. Mendonca, M.~E. Celebi, T. Mendonca, and J.~S. Marques.
\newblock Ph2: A public database for the analysis of dermoscopic images.
\newblock {\em Dermoscopy image analysis}, 2015.

\bibitem{menegola2017knowledge}
A. Menegola, M. Fornaciali, R. Pires, F.~V. Bittencourt, S. Avila, and E.
  Valle.
\newblock Knowledge transfer for melanoma screening with deep learning.
\newblock In {\em IEEE International Symposium on Biomedical Imaging}, pages
  297--300, 2017.

\bibitem{moss1996unsupervised}
R.~H. Moss, G.~A. Hance, S.~E. Umbaugh, and W.~V. Stoecker.
\newblock Unsupervised color image segmentation: with application to skin tumor
  borders.
\newblock 1996.

\bibitem{nachbar1994abcd}
F. Nachbar, W. Stolz, T. Merkle, A.~B. Cognetta, T. Vogt, M. Landthaler, P.
  Bilek, O. Braun-Falco, and G. Plewig.
\newblock The abcd rule of dermatoscopy: high prospective value in the
  diagnosis of doubtful melanocytic skin lesions.
\newblock {\em Journal of the American Academy of Dermatology}, 30(4):551--559,
  1994.

\bibitem{perez2018data}
F. Perez, C. Vasconcelos, S. Avila, and E. Valle.
\newblock Data augmentation for skin lesion analysis.
\newblock In {\em OR 2.0 Context-Aware Operating Theaters, Computer Assisted
  Robotic Endoscopy, Clinical Image-Based Procedures, and Skin Image Analysis},
  pages 303--311. Springer, 2018.

\bibitem{sivic2006video}
J. Sivic and A. Zisserman.
\newblock Video google: Efficient visual search of videos.
\newblock In {\em Toward category-level object recognition}, pages 127--144.
  Springer, 2006.

\bibitem{xue2018adversarial}
Y. Xue, T. Xu, and X. Huang.
\newblock Adversarial learning with multi-scale loss for skin lesion
  segmentation.
\newblock In {\em International Symposium on Biomedical Imaging}, pages
  859--863, 2018.

\end{thebibliography}
}

\end{document}